\pdfoutput=1

\documentclass[11pt]{article}

\usepackage[]{EMNLP2023}

\usepackage{times}
\usepackage{float}

\usepackage{latexsym} 
\newcommand{\Ar}[1]{{\scriptsize\<#1>}}
\usepackage{multirow}
\usepackage[T1]{fontenc}

\usepackage[utf8]{inputenc}

\usepackage{microtype}

\usepackage{inconsolata}

\usepackage{graphicx}
\usepackage{booktabs}
\usepackage{arabtex}
\usepackage{utf8}
\setcode{utf8}

\usepackage[colorinlistoftodos,prependcaption,textsize=tiny]{todonotes}

\newcommand{\idda}{\textsc{IDRISI-DA}}

\title{ArabicNLU 2024: The First Arabic Natural Language Understanding Shared Task}

\author{\normalsize \textbf{Mohammed Khalilia}$^{1}$ ~ \textbf{Sanad Malaysha}$^{1}$~   \textbf{Reem Suwaileh}$^{2}$ ~ \textbf{Mustafa Jarrar}$^{1}$\\
\normalsize \textbf{Alaa Aljabari}$^{1}$ ~ \textbf{Tamer Elsayed}$^{3}$ ~ \textbf{Imed Zitouni}$^{4}$ ~ \\
\normalsize $^{1}$Birzeit University, Palestine ~
  \normalsize $^{2}$Hamad Bin Khalifa University, Qatar\\
  \normalsize  $^{3}$Qatar University, Qatar ~
  \normalsize  $^{4}$Google, USA\\ %
  \texttt{\normalsize \{mkhalilia, smalaysha, mjarrar, aaljabari\} @birzeit.edu ~} \\ \texttt{ \normalsize rsuwaileh@hbku.edu.qa ~ \normalsize telsayed@qu.edu.qa ~ \normalsize imed.zitouni@gmail.com ~} 
}

\usepackage{abstract}
\begin{document}
\maketitle

\begin{abstract}
This paper presents an overview of the Arabic Natural Language Understanding (ArabicNLU 2024) shared task, focusing on two subtasks: Word Sense Disambiguation (WSD) and Location Mention Disambiguation (LMD). The task aimed to evaluate the ability of automated systems to resolve word ambiguity and identify locations mentioned in Arabic text. We provided participants with novel datasets, including a sense-annotated corpus for WSD, called SALMA with approximately $34$k annotated tokens, and the \idda{} dataset with $3,893$ annotations and $763$ unique location mentions. These are challenging tasks. Out of the $38$ registered teams, only three teams participated in the final evaluation phase, with the highest accuracy being $77.8\%$ for WSD and the highest MRR@1 being $95.0\%$ for LMD. The shared task not only facilitated the evaluation and comparison of different techniques, but also provided valuable insights and resources for the continued advancement of Arabic NLU technologies.
\end{abstract}

\section{Introduction}

Natural Language Understanding (NLU) is a core aspect of Natural Language Processing (NLP), facilitating semantics-based human-machine interactions \citep{BenderK20}. One of the key challenges in Arabic is ambiguity, because Arabic exhibits morphological richness, encompassing a complex interplay of roots, stems, and affixes, and rendering words susceptible to multiple interpretations based on their morphology \citep{J21}. Ambiguity in language can lead to misunderstandings, incorrect interpretations, and errors in NLP applications \citep{maulud2021state}. A core NLU task is Word Sense Disambiguation (WSD), and its special case Location Mention Disambiguation (LMD). WSD aims to determine the correct sense of ambiguous words in context \citep{JMHK23,HJ21b}, while LMD focuses on disambiguating location mentions that are referred to with multiple toponyms, i.e., particular place or location \citep{rsuwaileh2023idrisid}. 

\begin{figure}[ht!]
    \centering
\includegraphics[width=0.45\textwidth]{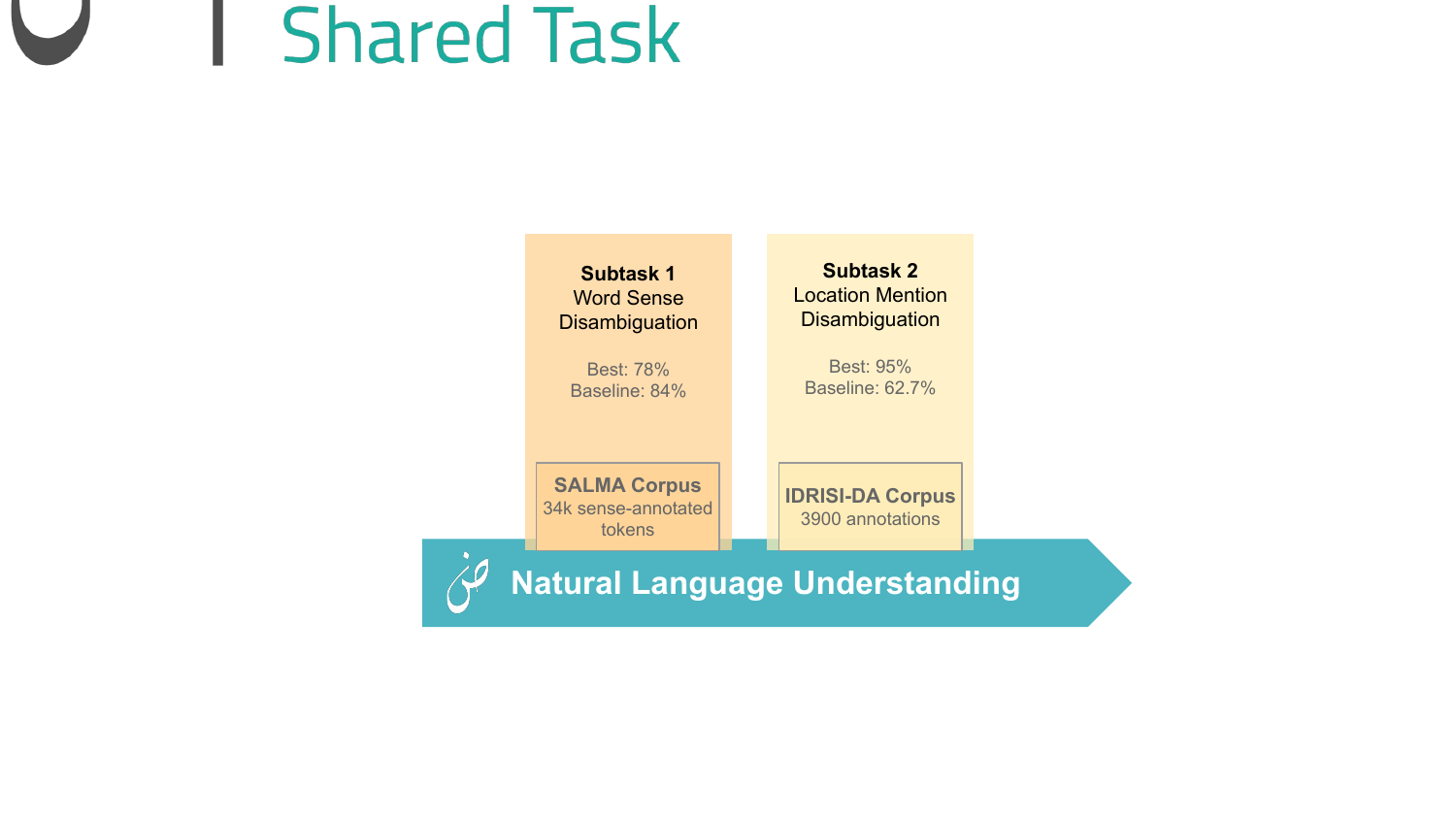}
    \caption{ArabicNLU datasets for WSD and LMD.}
    \label{fig:domains}
\end{figure}

The Arabic linguistic complexity, coupled with inherent polysemy, underscores the necessity for these lexical disambiguation tasks. They help decipher the intended sense of a word and the targeted entity of a mention within diverse contexts. For instance, in the sentence ``\Ar{شربت من عيون طرابلس وأكلت حلاوة الجبن فيها} / I drank from the springs of Tripoli and ate the sweetness of cheese in them'' one needs to disambiguate the meaning of two ambiguous words: ``\Ar{عيون} / springs,'' which has $11$ related senses, and ``\Ar{طرابلس} / Tripoli,'' which could refer to a location in either Lebanon or Libya. 

WSD is particularly important for tasks like machine translation \citep{RaganatoST20}, where it plays a pivotal role in improving accuracy by selecting contextually appropriate translations. Information retrieval systems \citep{wsdir} also heavily rely on accurate WSD and LMD to ensure search queries yield relevant results, considering the appropriate senses/entities of words within queries. Furthermore, applications such as question answering \citep{BakariBN21}, sentiment analysis \citep{sentimentanalysis}, text summarization \citep{summarizaion}, news analysis \citep{fakenews}, and semantic search \citep{semanticsearch} benefit from WSD and LMD. These tasks contribute to a nuanced understanding of Arabic text, enhancing the accuracy and relevance of results across diverse NLP applications. Recently, semantic disambiguation tasks have become integral to addressing hallucinations in Large Language Models (LLMs) \citep{hallucination,BCGJMSSV24}. 



While these tasks are extensively researched in well-resourced languages, there is a noticeable lack of focus on Arabic, despite their pivotal role in NLP \citep{MJK23}. This scarcity in Arabic NLP research can largely be attributed to the lack of datasets supporting these essential tasks \citep{JMHK23}. Without sufficient data, researchers face significant obstacles in developing and evaluating models tailored to the complexities of the Arabic language.

To address these challenges and draw attention to the issues faced in Arabic NLU, we organized 
the first Arabic Natural Language Understanding (ArabicNLU 2024) shared task, focusing on two subtasks: WSD and LMD. We have provided the participating teams with carefully annotated datasets, which are publicly accessible. 
Specifically, we have provided two manually annotated high-quality datasets for Arabic WSD and LMD. The teams were invited to experiment with diverse deep learning and machine learning methodologies, including, but not limited to, generative approaches, multi-task learning, transfer learning, sequence classification, sequence-to-sequence modeling, and graph models. Despite having $38$ registered teams, we received only three submissions, highlighting the challenging and non-trivial nature of the shared task topics.

The remainder of the paper is organized as follows: Section \ref{sec:related_works} offers a brief literature of Arabic WSD and LMD. Section \ref{sec:subtasks_description} details the intricacies of the shared task. Section~\ref{sec:subtask1_wsd} discusses the WSD task, including definition, dataset, baselines, participants’ systems, and results. Section~\ref{sec:subtask2_lmd} discusses the LMD task, including definition, dataset, baselines, participants’ systems, and results. Finally, Section \ref{sec:conclusion} concludes the paper.
\section{Related Work}
\label{sec:related_works}
NLU enables language models to accurately represent the knowledge embedded in words, which is crucial for core semantic tasks like WSD. WSD remains challenging despite the advancements in deep learning models like Bidirectional Encoder Representations from Transformers (BERT) and Generative Pre-trained Transformers (GPT). The challenge extends beyond WSD, involving various disambiguation tasks such as LMD, which aims at disambiguating multiple toponyms for a given location. More complex disambiguation tasks involve intents, anaphora, metaphors, and poetry. Therefore, computational semantics must explore these areas in greater depth, beyond merely considering the Zipfian distribution of words.


\subsection{Word Sense Disambiguation (WSD)} 
\paragraph{Systems} 
Traditionally, rule-based methods \citep{rulebased} dominated utterance ambiguity approaches by leveraging lexical resources such as Qabas \citep{JH24} and WordNet \citep{englishWordNet}. Later machine learning techniques such as Support Vector Machine (SVM) and Naive Bayes \citep{wsdmltechs} became predominant, employing supervised learning techniques on labeled datasets. More recently, the rise of deep learning has significantly advanced the field, with neural network models, such as Convolutional Neural Networks (CNNs) and Recurrent Neural Networks (RNNs) \citep{HJ21,wsddl}. Transformer-based encoder models like BERT and its variants \citep{kenton2019bert} have further revolutionized WSD by leveraging large-scale pre-training on extensive corpora, followed by fine-tuning on specific WSD adapted datasets \citep{MJK23}. Despite these advancements, Arabic NLP has seen slower progress in tackling WSD. To bridge this gap and stimulate further research, we present the ArabicNLU-2024 shared task. This initiative introduces a robust and rich datasets, aiming to propel Arabic NLU development and ultimately enhance human-computer interaction across diverse tasks.
\paragraph{Evaluation} 
While this task is often better studied in English due to the availability of extensive resources, Arabic lacks adequate datasets and knowledge bases, necessitating the curation of high-quality resources to advance the research and support the Arabic NLP community \citep{Elayeb19}. To address WSD, we have introduced the SALMA dataset \citep{JMHK23} as evaluation benchmark, which is a sense-annotated corpus with meanings extracted from two parallel lexicons: Al-Ghani Al-Zaher \citep{ghaniabul2014} and Contemporary Arabic Dictionary \citep{contemporaryomar2008}. This corpus comprises \textasciitilde$34K$ tokens, all annotated with their candidate meanings, considering the relatedness of each sense to the actual meaning of the word in context. Although other corpora have been designed for Arabic WSD, none fully meet the task's requirements. For instance, the Arabic version of the OntoNotes WSD dataset \citep{ontonotes}, annotated for three languages, lacks a well-defined sense list due to merging senses in a very coarse-grained manner. Similarly, the AQMAR dataset \citep{aqmar} for Arabic WSD was not annotate using senses, but instead utilized high-level lexical classes. Nonetheless, we considered F1-score for evaluating the systems on SALMA because such metric represents balanced view for both precision and recall.


\subsection{Location Mention Disambiguation (LMD)} 
\paragraph{Systems} 
A few studies have addressed the LMD task for English language using machine learning and deep learning techniques. For example, Geoparspy \cite{middleton2018} uses SVM trained on gazetteer-based features. Additionally, \citet{wang2019a} employed machine learning models for their toponym resolution system, including: (i) \textit{DM\_NLP}~\cite{wang2019dm}, a Light Gradient Boosting Machine (LightGBM), (ii) \textit{UniMelb}~\cite{li2019unimelb}, an SVM classifier, and (iii) \textit{UArizona}~\cite{yadav2019arizona}, a heuristic-based system that favors toponyms with higher populations. Furthermore, \citet{Xu2019DLocRL} proposed an attention-based model using two pairs of bi-LSTMs to match location mentions against the Foursquare gazetteer. 
The two-pair networks learn the left and right contexts of the LM, and both representations are processed through a fully connected layer for disambiguation.

\paragraph{Evaluation} 
There is a dearth of public LMD datasets. In this shared task, we use the \emph{only} public Arabic LMD dataset, \idda{} \cite{rsuwaileh2023idrisid}, for evaluation. 
Discrete metrics such as Accuracy (Acc), Precision (P), Recall (R), and F$_{\beta}$ scores are the most common metrics used to evaluate LMD systems~\cite{zhang2014geocoding,li2014effective,ji2016joint,middleton2018,wang2019a,Xu2019DLocRL}. However, these provide a broad overview and miss the nuances of different techniques. Distance based metrics assess LMD systems by measuring the great circle distance between the GPS coordinates of the gold and predicted location mentions, with overall performance computed by Median and Mean Error Distance. Acc, P, R, and F$_{\beta}$ can also be computed within a distance $d$, commonly set to $161$ km ($100$ miles). A significant issue with distance-based measures is the need to dynamically adjust the threshold for acceptable distance errors based on varying location granularity. While these measures are suitable for binary classification tasks, LMD is typically modeled as a multi-class classification or ranking task, making these measures less appropriate for evaluation. To address all these issues, we use Mean Reciprocal Rank at cutoff $k$ (MRR@k).


\subsection{Shared tasks}
The ArabicNLU shared task is the first to address both word and location disambiguation in Arabic, marking a significant milestone in the field. This initiative is supported by other notable shared tasks aimed at understanding Modern Standard Arabic (MSA) and dialects. These include FinNLP for financial text processing \cite{AraFinNLP24} using the \cite{JBKEG23} dataset, NADI for dialect identification \cite{NADI2023} based on the \cite{abdul-mageed-etal-2018-tweet} dataset, and WojoodNER for named entity recognition \cite{JHKTEA24, JAKBEHO23} utilizing the Wojood dataset \citep{JKG22}. Collectively, these collaborative efforts and interdisciplinary research projects foster a comprehensive understanding of linguistic nuances and enhance the applicability of NLP techniques across various contexts

\section{Shared-task Overview}
\label{sec:subtasks_description}
The ArabicNLU shared task consists of two primary sub-tasks, WSD and LMD. The WSD sub-task focuses on determining the correct semantic meaning of words (i.e., disambiguation of the word semantics) in a given context. The LMD sub-task, a special case of WSD, aims to accurately identify and disambiguate location mentions based on their geographical context.

The shared task mandates the use of pre-defined sense and location inventories that are directly linked to the provided datasets. Participants are prohibited from altering the senses, location mentions or toponyms within the test set. However, they are allowed to utilize external data and resources, including generative models, to improve their algorithms and models performance. To facilitate a unified evaluation, CodaLab,\footnote{{\scriptsize\url{https://codalab.lisn.upsaclay.fr/competitions/17758}}}\footnote{{\scriptsize\url{https://codalab.lisn.upsaclay.fr/competitions/18918}}} a well-established platform for scoring shared task submissions, was employed. Furthermore, to guarantee equitable access to task guidelines and data, a dedicated web page\footnote{{\scriptsize\url{https://sina.birzeit.edu/nlu\_sharedtask2024/}}} was established for the shared task, providing detailed information to all participants.

We received registrations from $38$ unique teams. During the testing phase, $4$ teams submitted a total of $40$ entries, among which $27$ for the WSD subtask, and $13$ for the LMD subtask. We received three description papers from the participated teams, all of which were accepted.  Table \ref{table:teams} provides a detailed overview of the participated teams in alphabetical order by their name, including their affiliations and the tasks they participated in.

\begin{table*}[ht!]
\centering
\begin{tabular}{l l l}
\hline
\textbf{Team} & \textbf{Affiliation} & \textbf{Task} \\ 
\hline
Pirates \cite{wael2024pirates} & Nile University & WSD \\
Rematchka \cite{abdel-salam2024rematchka}  & Cairo University & WSD, LMD \\
Upaya \cite{rajpoot2024upaya} & SCB DataX & WSD, LMD \\
\hline
\end{tabular}
\caption{Overview of participated teams and their tasks.}
\label{table:teams}
\end{table*}

\section{Subtask 1: Word Sense Disambiguation}\label{sec:subtask1_wsd}
\subsection{Task Definition}
\begin{figure}[ht!]
    \centering
    \includegraphics[scale=0.60]{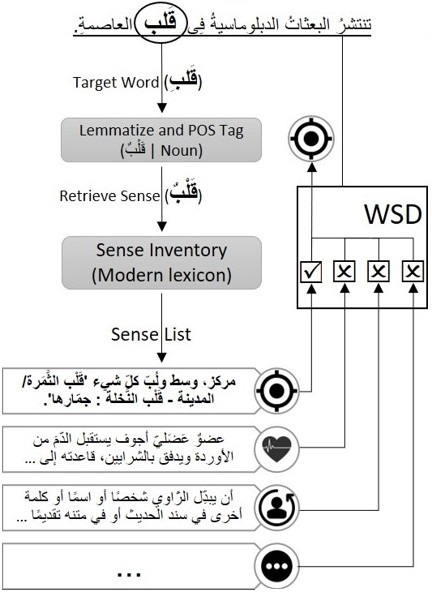}
    \caption{Illustration of the WSD subtask.}
    \label{fig:wsd}
\end{figure}

Polysemous words that convey multiple meanings in different contexts have led to the emergence of the WSD task \cite{J21}.
WSD aims to determine the intended semantic meaning (i.e., sense) of a word within a given context \cite{HJ21b, MJK23}. Given a context $c$ (i.e., a sentence), a target word $w$ in $c$, and a set of candidate senses $S = \{s_1, ..., s_n\}$, for the target word $w$, the goal of the WSD task is to determine which of these senses is the intended meaning of $w$. Figure \ref{fig:wsd} depicts the WSD sub-task.  

Participants were encouraged to utilize deep learning and generative methods for the WSD task. They were provided with a sense-annotated corpus of $1,340$ sentences, part of the SALMA corpus \cite{JMHK23}. Each target word in a given sentence has a set of candidate senses (glosses). Participants' submissions is expected to be in \textsc{JSON} format and must include the sentence id corresponrding to the context $c$, the word id of $w$, and the sense id of the sense $s_i$ (from the candidate senses $S$) for each target word in the test set. 

\subsection{Dataset: SALMA}


SALMA corpus \cite{JMHK23}, the first sense-annotated Arabic corpus, contains $1,440$ sentences and $34$k tokens, including $8,760$ unique tokens and $3,875 $ unique lemmas. Manually annotated with $4,151$ senses, it includes $19,030$ nouns, $2,763$ verbs, $7,116$ functional words, and $5,344$ punctuation marks and digits, as detailed in Table \ref{table:salma-corpus}. The data was collected from $33$ online media sources in Modern Standard Arabic (MSA), and has a $92\%$ inter-annotator agreement (IAA) measured using Quadratic Weighted Kappa.

In the shared task, participants received development and test sets. The development set includes $100$ sentences with corresponding candidate senses $S$ and the correct sense $s_i$ for each target word $w$ in a given sentence $c$. The remaining $1,340$ sentences were reserved for the test set, which included candidate senses, but excluded the correct sense. No training set was provided to encourage adoption and evaluation of generative model techniques, and participants were encouraged to use external datasets, sense inventories, or lexicons in their systems.

\begin{table}[ht]
\centering
\small
\setlength{\tabcolsep}{5pt} 
\begin{tabular}{@{}lcccc@{}}
\toprule
\textbf{Term} & \textbf{Nouns} & \textbf{Verbs} & \begin{tabular}[c]{@{}c@{}}\textbf{Func.}\\ \textbf{Words}\end{tabular} & \begin{tabular}[c]{@{}c@{}}\textbf{Punct.}\\ \textbf{\& Digits}\end{tabular} \\
\midrule
Total Tokens & $19,030$ & $2,763$ & $7,116$ & $5,344$ \\
Unique Tokens & $6,670$ & $1,593$ & $322$ & $175$ \\
Unique Lemmas & $2,904$ & $677$ & $119$ & $175$ \\
Unique Senses & $3,151$ & $792$ & $206$ & $2$ \\
\bottomrule
\end{tabular}
\caption{Statistics of SALMA corpus.}
\label{table:salma-corpus}
\end{table}

\subsection{Baselines}
Our WSD baseline approach involved developing a BERT-based system using Target Sense Verification (TSV) models. The TSV model is trained on a binary classification task that assigns confidence scores for True and False labels to each context-gloss pair. We created context-gloss pairs for each word in SALMA with varying context sizes to assess their impact on accuracy. The intended meaning was determined by ranking the glosses based on their True confidence scores, then selecting the one with the highest score as the intended gloss. Table \ref{table:wsd-baseline} presents our baseline model' performance using Accuracy across diverse context window sizes. For instance, a window size of $11$ encompassed five words on each side of the target word in the surrounding context, while full context refers to the entire sentence. Our best-performing WSD baseline model achieved an accuracy (F1-score) of $84.2\%$.

\begin{table}[H]
\centering
\begin{tabular}{ll}
\hline
\textbf{Context Window Size}      & \textbf{Baselines (F1-score)} \\ \hline
3    & $82.80\%$                  \\
5   & $84.00\%$                 \\
7 & $83.80\%$                  \\
9  & $83.50\%$                  \\
11  & $\textbf{84.20\%}$         \\
full     & $82.80\%$                   \\ \hline
\end{tabular}
\caption{Baselines of WSD.}
\label{table:wsd-baseline}
\end{table}

\subsection{Participants’ Systems}
Thirty five teams registered for the WSD subtask, out of which only three teams submitted their system descriptions as shown in Table \ref{table:teams}. Next we explore their approaches and results.

\noindent \textsc{\textbf{Upaya} \cite{rajpoot2024upaya}}:
They leveraged LLMs, specifically Llama3 \citep{llama3modelcard} and GPT-4 \citep{gpt4}, utilizing zero-shot learning techniques. The team employed a prompt-based approach where they manually crafted a natural language task description to be used consistently across all experiments. Initially, they experimented with a basic prompt that outputs plain text, then they enhanced the prompt by adding instructions that structures the input and output in JSON format to improve the model's comprehension. This structured format showed notable improvements with the Llama-3-70B-Instruct \citep{llama3modelcard} model. Additionally, they explored in-context learning by providing example sentences, target words and definitions in the prompt.

\noindent \textsc{\textbf{Pirates} \cite{wael2024pirates}}:
The approach used by this team involves leveraging transformer-based models, specifically focusing on AraBERTv2 \citep{arabertantoun2020}, through three main experiments: using Sentence Transformers with Siamese networks \citep{siamese}, the SetFit framework \citep{setfit2024}, and a classification approach. The first experiment involves fine-tuning AraBERTv2 as a Sentence Transformer with contrastive loss, where the model learns to differentiate between positive and negative senses of a word within a sentence by calculating the Euclidean distance between their embeddings. This method uses a combined dataset prepared by integrating two resources, Al-Ghani Al-Zaher lexicon \citep{ghaniabul2014} and Arabic Context Gloss pairs \citep{wsdrazzaz}, to ensure the model is exposed to both positive and hard negative samples. In the second experiment, they utilize the SetFit framework optimized for few-shot learning, which is advantageous due to its efficiency with minimal data input. This approach involves training the model on the sentence, target word, and its meaning, all separated by special tokens, and applying a cosine similarity loss function. The third experiment employs a more traditional classification approach using a transformer model for sequence classification. The AraBERTv2 model is fine-tuned with the SALMA development dataset, with the input structured similarly to the SetFit approach, but using the AdamW optimizer and training for fewer epochs. This method has shown the highest performance in terms of $F_1$-score among their three experiments.

\noindent \textsc{\textbf{Rematchka} \cite{abdel-salam2024rematchka}}:
The participants employed zero-shot learning using LLMs and fine-tuning of pre-trained language models (PLMs). They explored the effectiveness of different models such as LLama3, WizardLM-2 \citep{wizard2023}, AceGPT \citep{acegpt2023}, and OpenChat \citep{openchat2023}. In the zero-shot setting, the models were instructed to select the appropriate sense from a list of senses given the context and target word. This approach aimed to leverage the general language understanding capabilities of the models to perform WSD without task-specific training. Additionally, fine-tuning models like MARBERT \citep{marbert2021} and AraBERT \citep{arabertantoun2020} were explored to enhance performance in WSD tasks.

\subsection{Results}
Table \ref{table:wsd-results} summarizes the final results of the participating teams on the test dataset. The top-performing team, \textsc{Upaya}, achieved a $78\%$ accuracy with zero-shot learning technique using Llama3-70B-Instruct, outperforming GPT-4. In the second-place  \textsc{Pirates}, attained a $71\%$ accuracy by fine-tuning a sense classifier using AraBERTv2, a model proven effective for Arabic. In the third place is \textsc{Rematchka}, which employed multiple generative models for zero-shot learning, but their prompt-based approach yielded the lowest performance with a $56\%$. 

Notably, none of the participants surpassed our baseline ($84.2\%$). This may suggest that generative models utilized by the participants, specially in zero-shot settings, still fall short of outperforming an encoder-based model fine-tuned on a discriminative task using high-quality large dataset. Generative models are also limited in their multilingual support as the majority of their training data covers English language. For instance, only 5\% of the Llama3-70B training data is multilingual, covering 30 languages. It also an open question, whether the embeddings of causal generative models are less effective then bi-directional transformers for classification tasks.

\begin{table}[H]
\centering
\begin{tabular}{ll}
\hline
\textbf{Team} & \textbf{$F_1$-score} \\ \hline
Baseline      & \textbf{$84.2\%$}   \\ \hline
Upaya         & \textbf{$77.8\%$}     \\
Pirates       & $70.8$\%              \\
Rematchka     & $57.5$\%              \\ \hline
\end{tabular}
\caption{Results of participants on WSD subtask test data.}
\label{table:wsd-results}
\end{table}

\section{Subtask 2: Location Mention Disambiguation}
\label{sec:subtask2_lmd}
\subsection{Task Definition}
\begin{figure*}
    \centering
    \includegraphics[width=\linewidth]{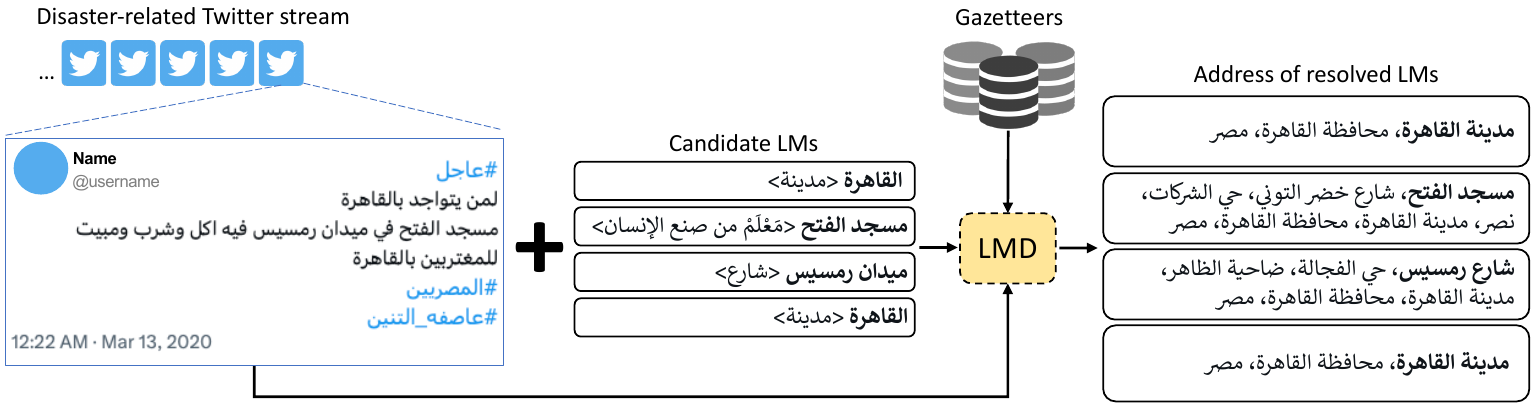}
    \caption{High-level overview of the LMD task.}
    \label{fig:lmd}
\end{figure*}

LMD represents a challenging problem in retrieval and classification, primarily due to issues such as the lack of context, toponymic polysemy, and toponymic homonymy \cite{rsuwaileh2023idrisid}. Figure~\ref{fig:lmd} presents a high-level overview of the task.

We formally define LMD problem as follows: Given a post $p$, the list of location mentions in $p$ $L_p=\{l_i: i \in [1, n_p]\}$, where $n_p$ is the number of location mentions in $p$, and a gazetteer $G=\{t_j: j \in [1, n_G]\}$, where $n_G$ is the number of toponyms in $G$, an LMD system aims to match every location mention $l_i$ in $p$ to a toponym $t_j$ in $G$ that accurately represents it, if exists. Otherwise, the system must abstain and declare that $l_i$ is unresolvable.

We perceive the LMD task as a candidate retrieval and ranking problem. For each location mention $l_i$, the LMD system must retrieve a ranked list of up to three candidate toponyms $R$ from OpenStreetMap (OSM), where $R \subset G$. Toponyms retrieved by $R$ are ranked based on the probability that each candidate is the correct toponym for $l_i$. Therefore, the LMD problem can be typically decomposed into two sub-problems: (i) candidate retrieval, which aims to retrieve a list of candidate toponyms from $G$, and (ii) candidate reranking, which aims to rerank the retrieved candidates $R$ in order of likelihood.


\subsection{Dataset: \idda{}}
\begin{figure}[!th]
    \centering
    \includegraphics[width=\linewidth]{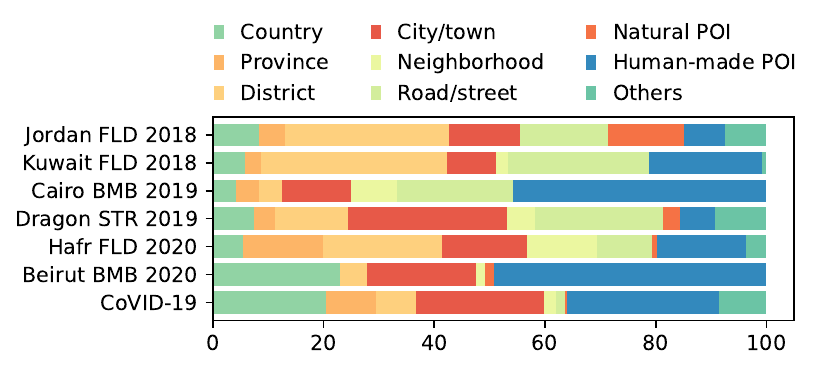}

    \label{fig:type_dist_uniq}
    \caption{Distribution of location types in \idda{}.}
    \label{fig:type_dist}
\end{figure}

The \idda{} dataset was created in two phases: extracting location mentions~\cite{rsuwaileh2023idrisira} and disambiguating them~\cite{rsuwaileh2023idrisid}. It is the first Arabic manually-labeled LMD dataset, designed with a particular attention on domain and geographical generalizability. Figure~\ref{fig:type_dist} shows the distribution of location types in \idda{}, per disaster event, showing its domain and geographical coverage, therefore exhibiting a reasonable dialectical coverage~\cite{rsuwaileh2023idrisira}. It includes 2,869 posts from X platform in diverse dialects, featuring 3,893 location mentions, with 763 unique mentions across seven countries. The dataset is split per event in ratios of 70:10:20 for training, development, and test sets, respectively. Each location mention in \idda{} is annotated with only one correct toponym extracted from OSM, containing attributes such as geo-coordinates, location type, and addresses, among others. 

\subsection{Baseline}
We compare the performance of the participated systems against \textsc{OSM}, a 
simple and common baseline for geolocation tasks. We specifically use Nominatim\footnote{https://nominatim.org} that runs over the official OSM online gazetteer.\footnote{\url{https://www.openstreetmap.org/}} 

\subsection{Participants’ Systems}
The LMD task had attracted $25$ registered teams, however, only $2$ of them managed to submit runs that we describe next.

\noindent \textsc{\textbf{Rematchka} \cite{abdel-salam2024rematchka}}: Used Llama3 \cite{llama3modelcard} to translate location mentions of type city or country to English if their type is verified. If not, Llama3 is queried for the most accurate country in English where the location mention is located, based on the post's context. The output is then passed to GeoPy\footnote{\url{https://geopy.readthedocs.io}} to retrieve corresponding toponyms. The rationale for translating to English is the GeoPy's degraded performance on Arabic text.

\noindent \textsc{\textbf{Upaya} \cite{rajpoot2024upaya}}: Proposed two retrieval stages approach. For every location mention, the system retrieves candidate toponyms from OSM, then re-ranks candidates using Cohere rerank-multilingual-v2.0.\footnote{\url{https://cohere.com/blog/rerank}} The Cohere reranker involves self-attention mechanisms and transformer-based architectures that capture the similarity between location mentions and candidate toponyms.

\subsection{Results}
We present the \emph{MRR@k} results of the participated systems in Table~\ref{tab:lmd_results}. The results demonstrate that both participants' systems outperform the baselines in \emph{MRR@1}, highlighting their effectiveness in retrieving the correct toponym from OSM at the top rank. Notably, the \textsc{Rematchka} system substantially outperforms all other systems across all measures, exhibiting superior performance. This indicates the robustness of GeoPy, particularly when used with the English language. These results underscore the need for developing more robust models for Arabic LMD.
\begin{table}[H]
    \centering
    \setlength{\tabcolsep}{5pt} 
    \begin{tabular}{l r r r} \toprule
    Team & MRR@1 & MRR@2 & MRR@3 \\\hline
    \textsc{OSM}$_{baseline}$ & 0.5724 & 0.6396 & 0.6428 \\ \hline
    \textsc{Rematchka} & \textbf{0.9497} & \textbf{0.9500} & \textbf{0.9500} \\
    \textsc{Upaya} & 0.5994 & 0.5994 & 0.5994 \\\bottomrule
    \end{tabular}
    \caption{Results of LMD participants on test set.}
    \label{tab:lmd_results}
\end{table}


\section{Conclusion and Future work}
\label{sec:conclusion}

In this paper, we present the results of the ArabicNLU 2024 shared task, focusing on the challenges of Word Sense Disambiguation and Location Mention Disambiguation in Arabic Natural Language Understanding. The findings from participated teams highlight the ongoing challenges and research gaps associated with these subtasks. We observe that generative models underperform traditional classification architectures trained on labeled data. This is specially the case for low resourced languages which are not well supported in LLMs. LLMs are mostly English-centric due to the imbalanced training corpora. This also extends to other systems and tools such as GeoPy, requiring machine translation to English to achieve the desired performance. The challenge of multilingual support becomes even more apparent when working with Arabic dialectical data. To really democratize Arabic NLP, it is essential to compile large datasets in various Arabic dialects, as demonstrated by the work of \cite{JZHNW23, EJHZ22, JZHNW23}. Additionally, new techniques must be developed to address the scarcity of dialectical data, and LLMs specifically tailored for the Arabic language need to be trained.

Our vision for this shared task is to create a collaborative environment that accelerates research and development in Arabic NLU. By facilitating the evaluation and comparison of various models and techniques, we aim to uncover new insights, foster innovation, and build a strong foundation of resources. This effort seeks to overcome current obstacles and significantly advance the capabilities of Arabic NLU technologies.
\section*{Limitations}

Acknowledging the inherent constraints within the ArabicNLU shared task datasets is crucial. The SALMA dataset, utilized for the WSD subtask, primarily employs an extended version of "Modern" as a sense inventory, which includes referral glosses that need specific handling for broader applicability. Furthermore, it is limited to MSA, excluding dialects, and focuses only on single-word lemma senses.

The \idda{} dataset, being crawled from X platform, faces significant limitation in its application due to recent X platform API restrictions that may reduce its utility for research focused on social media platforms. However, the dataset facilitates developing LMD systems the process informal text sourced from various platforms beyond X platform. Furthermore, fine-grained locations and temporary locations are underrepresented in \idda{}, those are pivotal during emergencies. Nevertheless, the goal of this shared task is to develop generic LMD systems not domain(disaster)-specific ones.
\section*{Ethics Statement}\label{sec:Ethics} 

The datasets provided for this shared task are derived from public sources, eliminating specific privacy concerns. The results of the shared task will be made publicly available to enable the research community to build upon them for the public good and peaceful purposes. Our data and techniques are strictly intended for non-malicious, peaceful, and non-military purposes.

\section*{Acknowledgements}
This research is partially funded by the research committee at Birzeit University. We extend our gratitude to Taymaa Hammouda for the technical support. The contribution of Reem Suwaileh is partially funded by the NPRP grant 14C-0916-210015, which is provided by the Qatar National Research Fund part of Qatar Research Development and Innovation Council (QRDI).

\bibliography{custom,MyReferences}
\bibliographystyle{acl_natbib}

\end{document}